\documentclass[runningheads]{llncs}

\usepackage{eccv}

\usepackage{eccvabbrv}

\usepackage{graphicx}
\usepackage{booktabs}
\usepackage{multirow}
\usepackage[ruled,linesnumbered, algo2e]{algorithm2e}
\usepackage{algorithm}
\usepackage{algorithmic}
\usepackage{tabularx}
\usepackage{appendix}
\usepackage[accsupp]{axessibility}  % Improves PDF readability for those with disabilities.

\usepackage{hyperref}

\usepackage{orcidlink}

\begin{document}

% ---------------------------------------------------------------
% TODO REVIEW: Replace with your title
\title{E4C: Enhance Editability for Text-Based Image Editing by Harnessing Efficient CLIP Guidance} 

% TODO REVIEW: If the paper title is too long for the running head, you can set
% an abbreviated paper title here. If not, comment out.
% \titlerunning{Abbreviated paper title}

% TODO FINAL: Replace with your author list. 
% Include the authors' OCRID for the camera-ready version, if at all possible.
\author{Tianrui Huang\inst{1}$^{,\dagger}$
Pu Cao\inst{1}$^{,\dagger}$ \and
Lu Yang\inst{1} \and
Chun Liu\inst{1} \and
Mengjie Hu\inst{1} \and
Zhiwei Liu\inst{2} \and
Qing Song\inst{1}$^{,*}$
}

% TODO FINAL: Replace with an abbreviated list of authors.
\authorrunning{T. Huang \& P. Cao et al.}
% First names are abbreviated in the running head.
% If there are more than two authors, 'et al.' is used.
\titlerunning{E4C}
% TODO FINAL: Replace with your institution list.
\institute{Beijing University of Posts and Telecommunications \and 
Institute of Automation Chinese Academy of Sciences\\
\email{\{huangtianrui, caopu, soeaver, chun.liu, mengjie.hu, priv\}@bupt.edu.cn  zhiweiliu@nlpr.ia.ac.cn}
}

\maketitle
\def\thefootnote{$\dagger$}\footnotetext{Authors contribute equally.}
\def\thefootnote{\arabic{footnote}}
\def\thefootnote{*}\footnotetext{Corresponding author.}
\def\thefootnote{\arabic{footnote}}

\newcommand{\RomanNumeral}[1]{\uppercase\expandafter{\romannumeral#1}}

\begin{abstract}
Diffusion-based image editing is a composite process of preserving the source image content and generating new content or applying modifications. While current editing approaches have made improvements under text guidance, most of them have only focused on preserving the information of the input image, disregarding the importance of editability and alignment to the target prompt. In this paper, we prioritize the editability by proposing a zero-shot image editing method, named \textbf{E}nhance \textbf{E}ditability for text-based image \textbf{E}diting via \textbf{E}fficient \textbf{C}LIP guidance (\textbf{E4C}), which only requires inference-stage optimization to explicitly enhance the edibility and text alignment. Specifically, we develop a unified dual-branch feature-sharing pipeline that enables the preservation of the structure or texture of the source image while allowing the other to be adapted based on the editing task. We further integrate CLIP guidance into our pipeline by utilizing our novel random-gateway optimization mechanism to efficiently enhance the semantic alignment with the target prompt. Comprehensive quantitative and qualitative experiments demonstrate that our method effectively resolves the text alignment issues prevalent in existing methods while maintaining the fidelity to the source image, and performs well across a wide range of editing tasks.
\end{abstract}

\section{Introduction}
\label{sec:intro}

Large-scale pre-trained text-to-image diffusion models\cite{rombach2022high, ramesh2022hierarchical, saharia2022photorealistic, nichol2021glide}, such as Stable Diffusion\cite{rombach2022high}, DALL-E2\cite{ramesh2022hierarchical}, and Imagen\cite{saharia2022photorealistic}, have demonstrated remarkable proficiency in text-to-image generation\cite{ruiz2023dreambooth, gal2022image, kumari2023multi}. They have even shown potential in more traditional tasks\cite{yang2023deep, yang2021attacks, yang2022quality}. More recently, research\cite{hertz2022prompt, tumanyan2023plug, couairon2022diffedit, kawar2023imagic, cao2023masactrl, parmar2023zero} has extended these diffusion models to image editing tasks. Despite the notable advances, however, we have identified two key challenges that may impede users from effectively utilizing existing methods.

\begin{figure}[t!]
    \centering
    \includegraphics[width=1\textwidth]{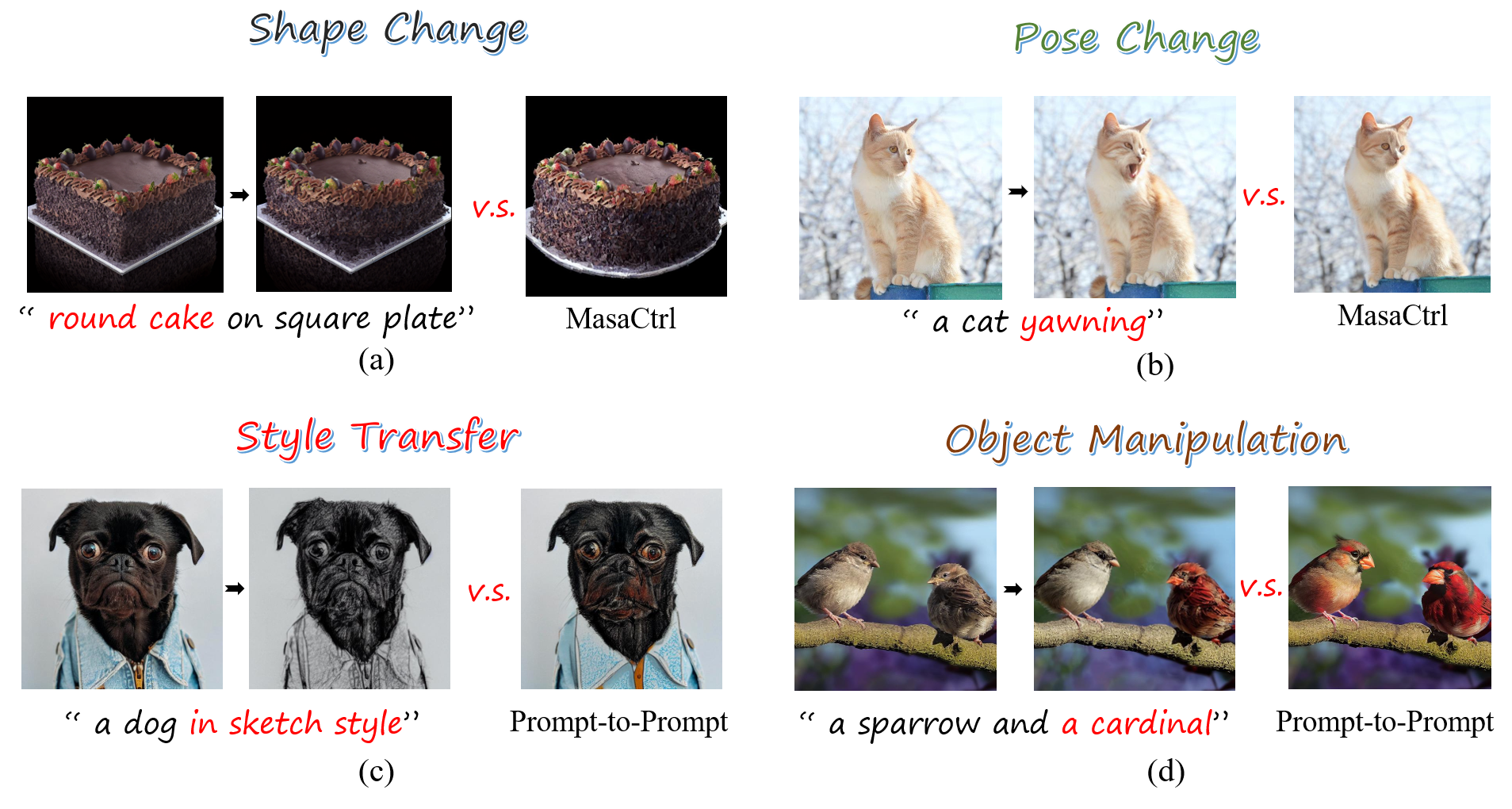}
    \caption{\textbf{E4C performs edits on various tasks.} Given a real image and a target text prompt, our method can generate a new image with high alignment to the description with no affiliation of masks or segmentation maps. Our method outperforms task-specif methods even in their advantageous domains.}
    \label{fig:teaser}
\vspace{-4mm}
\end{figure}

Firstly, existing methods\cite{hertz2022prompt, tumanyan2023plug, cao2023masactrl} have typically faced difficulties when tackling multiple types of editing tasks. Common tasks can be categorized into two main types: structure-consistent editing, which includes object replacement, attribute manipulation, etc., and non-rigid editing, involving changes to subject poses and shapes. Although some existing approaches have demonstrated success in dealing with one of these categories, they have generally struggled to handle the other. For instance, pioneering editing methods like prompt-to-prompt\cite{hertz2022prompt}, Blended-LD\cite{avrahami2023blended}, and Pix2pix-zero\cite{parmar2023zero} are capable of handling object attribute manipulation and replacement, but they struggle with structural modifications. Plug-and-Play\cite{tumanyan2023plug} focuses primarily on overall image style translation. MasaCtrl\cite{cao2023masactrl} is specifically designed for non-rigid editing but it falls short in common textual and content editing tasks. Although Imagic\cite{kawar2023imagic} can cater to more flexible editing requirements, it requires significantly more resources to train the entire U-Net\cite{ronneberger2015u} to memorize the key information of the source image and may run into dramatic background changes.

Secondly, revisiting existing editing approaches, we observe that the majority of the improvements have been made for preserving source information rather than aligning new content to the desired prompt. Prompt-to-prompt\cite{dong2023prompt} preservation of the cross-attention map from the source branch can help fix the object structure. Null-text inversion\cite{mokady2023null} and prompt tuning inversion\cite{dong2023prompt} tune the negative and positive text embeddings to align the latents of sampling process to the pivotal inversion track, rectifying the error raised by the naive DDIM inversion\cite{song2020denoising}. Plug-and-Play\cite{tumanyan2023plug} and MasaCtrl\cite{cao2023masactrl} leverage internal spatial features (feature maps and attentions in U-Net) to preserve the layout and appearance of the original image, respectively. While these methods ensure fidelity to the input image, they put less effort into explicitly improving the editability and text alignment. Here we refer \textit{editability} to the ability to generate new content that could be the transition of local attributes or the overall style. The Fig. \ref{fig:teaser} shows that existing methods\cite{hertz2022prompt}, in some cases, fail to generate the content according to text instructions: the opened mouth of the cat in (b) and the uncolored sketch style in (c). The \textit{text-alignment} represents how well the edited image matches the target prompt, failure cases are as (a) and (d) in Fig. \ref{fig:teaser}, where MasaCtrl\cite{cao2023masactrl}, though generate the round shape of cake, it changed the shape of bottom plate thus mistaking the text semantics, also there are two sparrows swapped by cardinals at the same time by P2P\cite{hertz2022prompt}.

We compose our method from two aspects to address these two main challenges. (\RomanNumeral{1}) To handle both consistent structure and complex non-rigid editing tasks, we consider that two types of editing tasks have different preservation needs: the structure-preserved tasks require maintaining the scene layout and object structure whereas selectively changing the texture and appearance to certain areas, which pattern is exactly contrast to the non-rigid tasks. Building on this insight, we devise our dual-branch feature-sharing pipeline to keep fidelity to the source image. More specifically, given that previous works\cite{tumanyan2023plug, cao2023masactrl, alaluf2023cross} have explored the role of intermediate features (especially from attention layers) of denoising U-Net in forming the fine-grained spatial and texture information, our pipeline shares between inversion and editing branches, either the queries in certain self-attention layers to preserve the overall structure and layout or keys-values for subject identity and texture. (\RomanNumeral{2}) To gain more text alignment upon the devised pipeline, we further utilize the CLIP model\cite{radford2021learning} as a supervisor to explicitly enhance the text alignment. To alleviate the resource burden when optimizing the iterative diffusion process, We design the random-gateway mechanism to facilitate the optimization process using CLIP guidance. We have experimentally demonstrated that our method can serve as a general pipeline that handles various editing tasks including both consistent structure and complex non-rigid types, while achieving high alignment to target prompt.

Our contributions can be summarized as follows:
\begin{itemize}
    \item[$\bullet$] For maintaining \textit{fidelity}, we construct a feature-sharing pipeline that enables adaptive preservation of information from source images so that it can flexibly cater to multiple types of editing tasks.
    
    \item[$\bullet$] For enhancing \textit{editability} and text alignment, we devise the random-gateway optimization mechanism to efficiently leverage the guidance from the CLIP model to rectify the editing process.

    \item[$\bullet$] Qualitative and quantitative results have shown that our method outperforms existing methods even in their advantageous domains and presents inherent superiority in dealing with hard samples.
\end{itemize}

\begin{figure}[t!]
\centering
\includegraphics[width=1\textwidth]{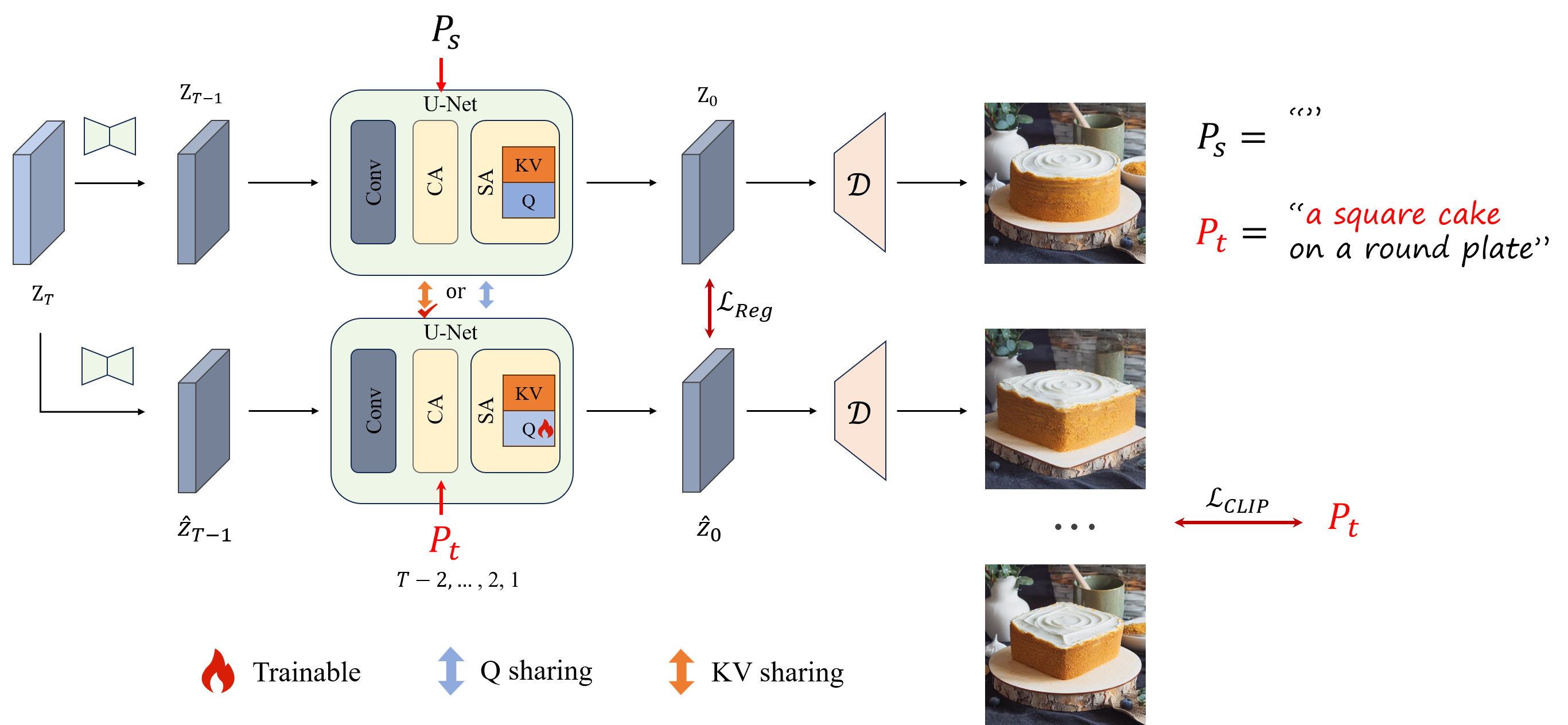}
\caption{\textbf{Dual-branch pipeline inserted with CLIP guidance.} We are attempting to change the shape of a cake from round to square. Thanks to the shared key-value pairs we maintain its appearance and surroundings. We further use CLIP loss (Eqn. \ref{eq: clip_loss}) to refine the Q features. By doing this, we can ensure that our final results stay consistent with $P_t$.}
\label{fig:dual_branch_pipeline}
\vspace{-4mm}
\end{figure}

\section{Related Work}

\subsection{Text-Guided Image Editing with Generative Models} 
There has been emerging research on manipulating images with generative models since the era of GAN's models\cite{goodfellow2014generative, cao2024decreases, cao2022lsap}. As the most widely used modality, research on image editing using natural language as instructions has surged. Methods like TAGAN\cite{nam2018text} and ManiGAN\cite{li2020manigan}, utilize text-conditional generators to manipulate images. However, they are restricted to specific domains and limited in the text variety, usually the classes of objects or attributes. Recently diffusion models\cite{nichol2021glide, rombach2022high}, which directly train the image generation tasks conditioned on text, have been demonstrated to be a more powerful open-world generator that can generate images with high quality and diversity. They have provided researchers with more possibilities for manipulating the image. Except for Blended Diffusion\cite{avrahami2022blended} and Blended LD\cite{avrahami2023blended} which perform edits by generating foreground objects with mask conditions and then blending them into the background, most diffusion-editing methods are inversion-based\cite{song2020denoising}. They first utilize the DDIM inversion\cite{song2020denoising} to get the noise latent and start the sampling process with it. Exemplary works\cite{hertz2022prompt, tumanyan2023plug, cao2023masactrl} like Prompt-to-prompt\cite{hertz2022prompt} injects the cross-attention map from the reconstruction branch into the editing one to preserve the original layout and use the new prompt to generate the content. Also, some works\cite{mokady2023null, dong2023prompt} are dedicated to improving the accuracy of DDIM inversion to get better reconstruction and edit results.

\subsection{Leverage of the CLIP Model for Image Manipulation}
Apart from the aforementioned approaches, there have also been several works that leverage the CLIP model\cite{radford2021learning} for image manipulation. It is because the CLIP model inherently aligns the vision and language modality which perfectly matches the goal of image editing under text guidance. StyleCLIP\cite{patashnik2021styleclip}, based on StyleGAN2\cite{gal2021stylegan}, manipulates the GAN's latent code to edit the image in a few specific domains. VQGAN-CLIP\cite{crowson2022vqgan} combines the VQGAN\cite{esser2021taming} and CLIP\cite{radford2021learning} to produce higher visual quality outputs. Text2Live\cite{bar2022text2live} trains an edit layer with CLIP loss to perform localized edits onto images. Most related to our work is DIffusionCLIP\cite{kim2022diffusionclip}, which also integrates CLIP into the diffusion process to get edit results with higher text alignment. However, our method has distinct differences with it: DiffusionCLIP is based on domain-specific diffusion models that process images in RGB space and our method is built upon a larger open-vocabulary latent diffusion model that samples in latent space encoded by VAE\cite{kingma2013auto}. Moreover, DiffusionCLIP directly trains the entire denoising U-Net at each timestep for in-domain global image manipulation and inter-domain transition while our method tweaks the high-level features at several timesteps to handle either micro details or the overall style, as well as non-rigid changes that DiffusionCLIP can barely deal with.

\begin{figure}[t!]
  \centering
  \includegraphics[width=1\textwidth]{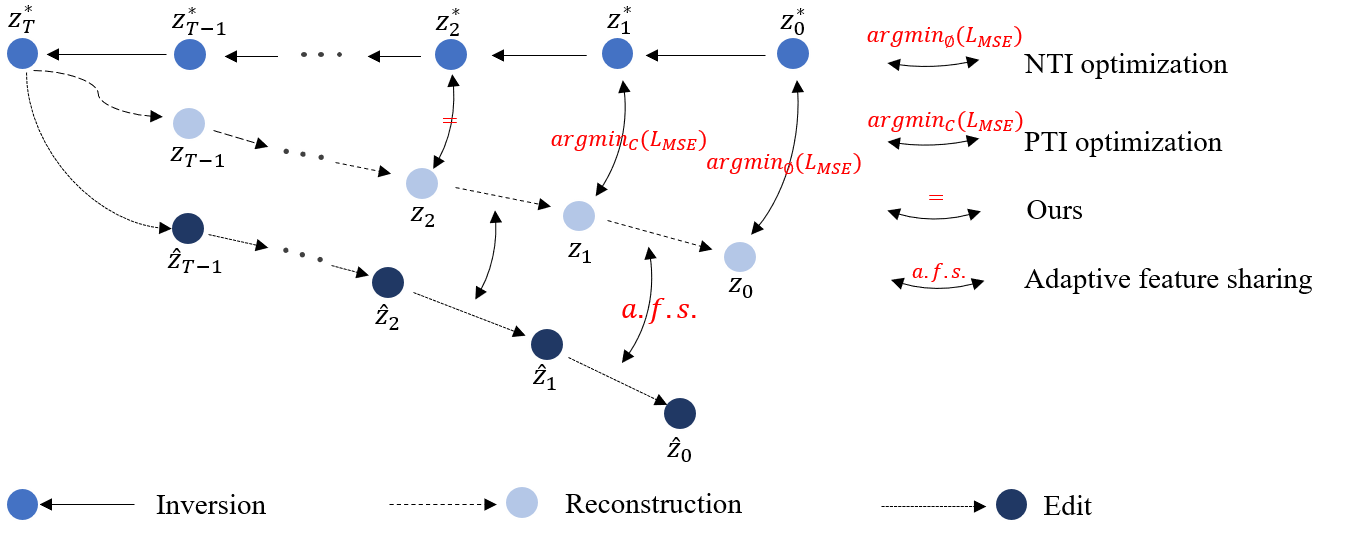}
  \caption{\textbf{High-level overview of our framework and comparison to previous methods.} From top to bottom lay our three branches: inversion, reconstruction (source), and editing (target). See the top two branches, We take different strategies of latent-align ($z_t$ to $z_t^*$) compared to NTI\cite{mokady2023null} and PTI\cite{dong2023prompt}. Besides, we apply adaptive feature sharing (a.f.s) between source and target branches. Note that all operations denoted by bidirectional curved arrow $\leftrightarrow$ are applied at each timestep.}
  \label{fig:dot_framework}
  \vspace{-3mm}
\end{figure}

\section{Method}
In the context of real image editing, we aim to manipulate certain parts of it under the guidance of a target prompt. Aimed editing tasks can include object-level replacement, attribute manipulations, overall style transfer, and more complicated non-rigid changes like altering on poses of creatures or shapes of items. Our terminal goal is divided into two aspects, to deal with all these types of edits through a single framework, and to guarantee a good alignment between the edit result and target text prompts. To this end, we propose the following key components to make our method:
\begin{enumerate}
    \item \textit{Faithful inversion feature preservation.} Naive DDIM inversion\cite{song2020denoising} suffers from inconsistency between inversion and sampling process which leads to bad reconstruction results. We propose that simple re-utilization of the intermediate U-Net features can perform much better reconstruction productivity and thus enhance the foundation of the editing process.
    \item \textit{Adaptive feature sharing.} We devise an adaptive feature-sharing gate to facilitate the interaction between the reconstruction and editing process, allowing us to effectively preserve key information from the source image unchanged.
    \item \textit{Efficient CLIP Guidance.} To leverage the pre-trained CLIP model\cite{radford2021learning} as a semantic supervisor, we design the random-gateway optimization mechanism to alleviate the memory burden during the iterative sampling process to optimize the key features in the noise predictor, thus achieving high alignment to the target prompt.
\end{enumerate}

\subsection{Faithful Inversion Feature Preservation}
\label{sec: faithful inversion}
For the general overview of this section, see Fig. \ref{fig:dot_framework}. Let $I$ be the input image, we first encode it to $z_0^*$ with VAE\cite{kingma2013auto} and then follow the reversed DDIM\cite{song2020denoising} sampling process to get the noise latent $z_T^*$. It is believed that $z_T$ contains encoded information of the source image and assembles the Gaussian samples, thus serving as the start code to the new generation process. Naive DDIM inversion can supply a pivotal trajectory of latents, denoted as $\{z_t^*\}_{t=0}^T$, and then it starts from the last noised latent $z_T=z_T^*$ and iteratively follows the Eqn. \ref{eq:sample} to denoise the latent. However, it has been shown that such a sapling process can barely produce satisfying reconstruction, especially when classifier-free guidance \cite{ho2022classifier} (cfg) is applied. A time-accumulated error leads to the collapse of the final results and impedes the fidelity of the source image.
\begin{align}
z_{t-1} = \sqrt{\frac{\alpha_t-1}{\alpha_t}}z_t + \sqrt{\alpha_t - 1}\left( \sqrt{\frac{1}{\alpha_t-1} - 1} - \sqrt{\frac{1}{\alpha_t} - 1} \right)\epsilon_\theta(z_t, t, C, \phi)
\label{eq:sample} 
\end{align}

Previous works have tried to optimize certain terms that can contribute to latent iteration, aiming to alleviate the deviation between inversion and the sampling process. As illustrated in Fig. \ref{fig:dot_framework}, Null-text inversion\cite{mokady2023null} (NTI) and Prompt Tuning Inversion\cite{dong2023prompt} (PTI) fine-tune the negative and positive text embedding, respectively at each timestep, to approximate $z_t$s to $z_t^*$s by minimizing the pair-wise MSE loss, thus help fixing the reconstruction collapse.
Unlike previous research, we directly substitute each $z_t$ with corresponding $z_t^*$ and feed it to U-Net. By doing this, we naturally reach the upper bound of the optimizing-based approach in the reconstruction process. On the other hand, we can faithfully reproduce the intermediate features as U-Net forwarding, by fixing the other input conditions in the sampling process as follows:
\begin{align}
\epsilon_\theta(z_t, t, C, \phi) &= \epsilon_\theta(z_t^*, t, C, \phi), \\
f_U &= f_U^*
\label{eq:faithful inversion}
\end{align}

To this extent, our inversion feature preservation strategy explicitly avoids the deviation between sampling and inversion process. At the reconstruction stage, we introduce no extra computing resources compared to optimization-based inversion methods, sparing more resources for later editability foresting.

\begin{figure}[t!]
\centering
\includegraphics[width=1\textwidth]{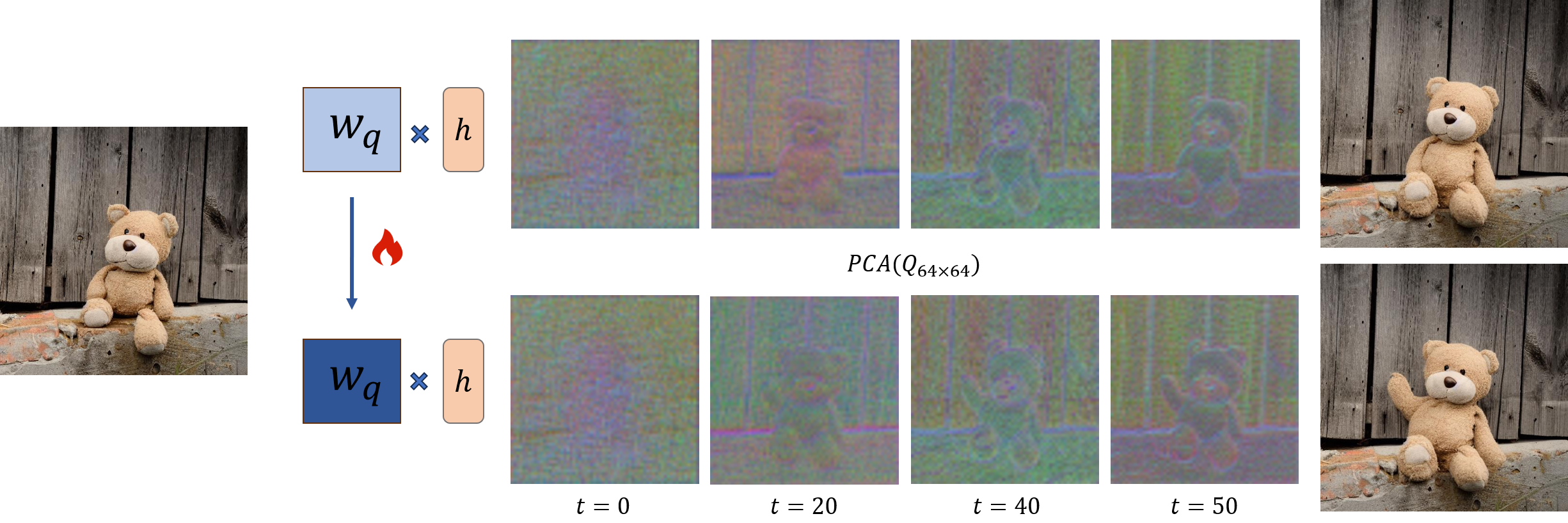}
\caption{\textbf{Visualizing queries before and after CLIP guidance.} We use PCA to visualize "making a teddy bear raise left hand" queries before and after fine-tuning with CLIP guidance, extracted from the 15th self-attention layer (16 in total).}
\label{fig:pca}
\vspace{-4mm}
\end{figure}

\subsection{Adaptive Feature Sharing}

Due to the intervention of the target prompt that can be semantically different from the input image, directly generating edit results with the new prompt can cause unintended errors. Recent research\cite{cao2023masactrl, alaluf2023cross} has shown that query, key, and value features of certain self-attention layers in denoising U-Net semantically capture the two aspects of information of the image: queries for structure and layout (see Fig. \ref{fig:pca}), whereas keys and values for textures and appearance, respectively. Our feature-sharing scheme is designed to cater adaptively to both structure-preserved and non-rigid image editing tasks. In detail, we transform the self-attention operation in the editing branch as follows:
\begin{equation}
\begin{aligned}
attn_t := 
    \begin{cases}
  \text{Softmax}\left(\frac{Q_sK_t^T}{\sqrt{d}}\right)V_t, & \text{for structure-preserved editing}, \\
  \text{Softmax}\left(\frac{Q_tK_s^T}{\sqrt{d}}\right)V_s, & \text{for non-rigid editing}.
    \end{cases}
\end{aligned}
\end{equation}
where subscript $s$ and $t$ denote the source and target branches (the reconstruction and editing branches mentioned in Fig. \ref{fig:dot_framework}), respectively. $d$ is the dimensionality of K. Note that the transformed attention is only applied to the editing branch and all operations in the source branch and is not affected.

Because the faithful inversion (Eqn. \ref{eq:faithful inversion}) gives a great reconstruction of the source branch, we can simply replace the queries or key-value pairs in the editing branch with those of the source branch to keep the fidelity to the source image. Say we want to replace a creature in a source image with another or alter some attributes of it, we need to keep the pose of the subject and the overall layout while tweaking part of the content and textures, then we are in a structure-preserved case and switch the pipeline to Q-sharing mode. And if we try to change a sitting dong into a running one or alter the shape of a cake from square to round, then the key-value feature pairs are shared between dual branches to keep the identity and appearance of the dog or the cake.

Our adaptive feature-sharing scheme is designed to preserve the essential information from the source in a flexible manner, based on the task at hand. This pipeline is capable of generating edit results that maintain the key features of the original input. However, there is no guarantee of proper alignment with the target prompt (Fig. \ref{fig:pca}). In the following section, we will introduce our efficient CLIP guidance to address this issue.

\begin{figure}[t!]
\centering
\includegraphics[width=1\textwidth]{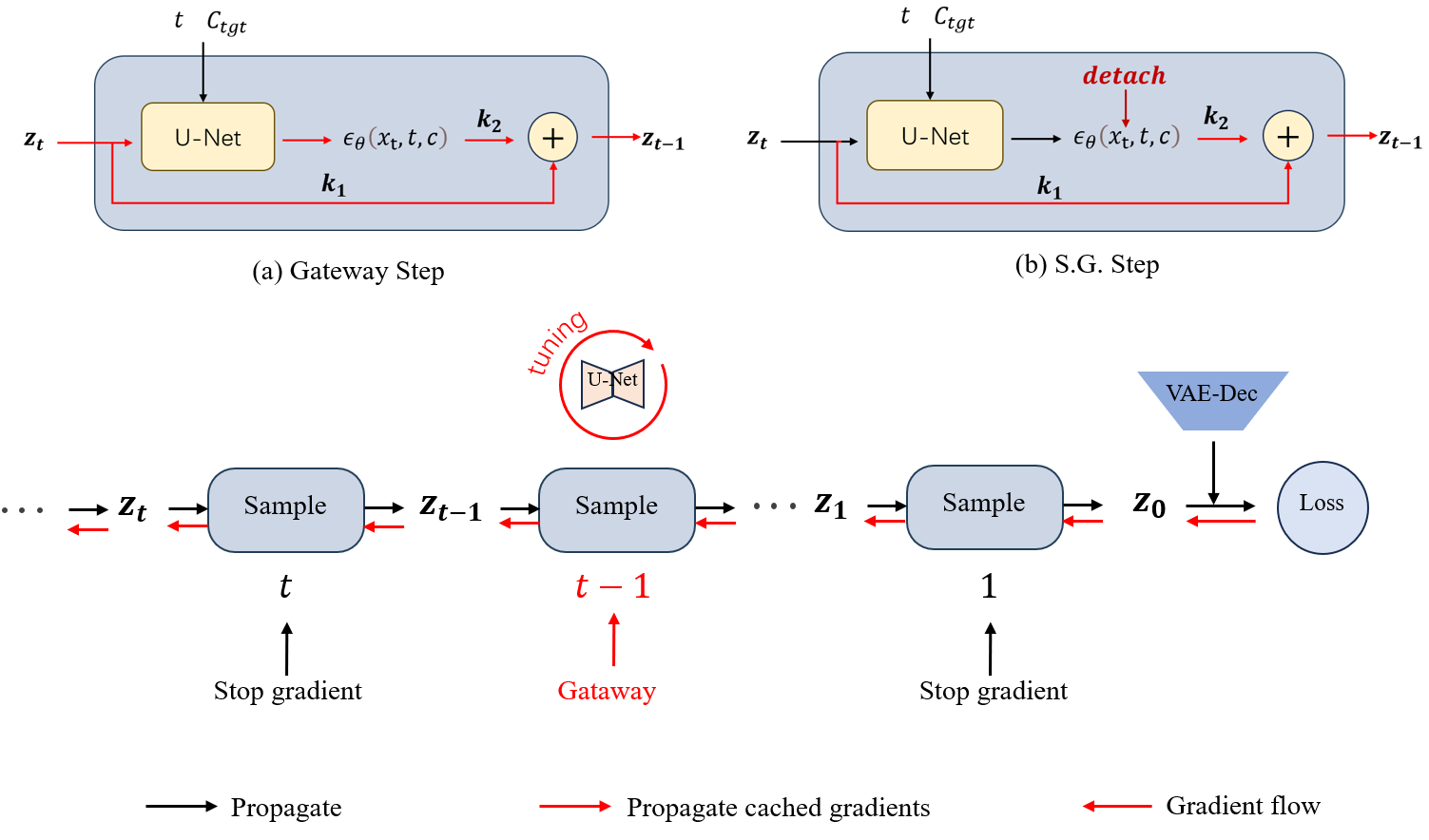}
\caption{\textbf{The illustration of our random-gateway optimization mechanism.} \textit{Top:} the computing graphics of one single sampling step. For a \textit{gateway} step, we cache all gradients contributed to U-Net parameters from both the noise term {$\epsilon_\theta$} and the residual latent term {$z_t$}. The noise term is stopped gradient at other steps, leaving only the latent path for the gradient signal passed to the previous step. \textit{Bottom:} The overview of gradient flow in the sampling process. The U-Net parameter would only be updated at the \textit{gateways} but the gradient flows through along the entire sampling process.}
\label{fig:random-gateway}
\vspace{-2mm}
\end{figure}

\subsection{Efficient CLIP Optimization}
As illustrated in Fig. \ref{fig:dual_branch_pipeline}, we compute CLIP loss\cite{radford2021learning} between the target prompt and the rough edit result output by our feature-sharing pipeline and then the gradient back-propagate through the editing process to update the U-Net. This process loops to moderately guide the edit result getting close to the target prompt semantically. 

Say we are dealing with the "making a teddy bear raise its left hand" task (Fig. \ref{fig:pca}), for instance, we need to keep the shared key-value pairs unchanged to maintain the teddy appearance and tune the queries for its pose-changing. We set the linear projection matrix in computing queries as optimizing parameters. Now in a shared self-attention layer, features in the source and editing branches are computed as follows:
\begin{equation}
\begin{aligned}
Q_s = W_qx, K_s = W_kx, V_s = W_vx \\
Q_t = W_q^tx, K_t = K_s, V_t = V_s
\end{aligned}
\end{equation}
where $W_q$, $W_k$, and $W_v$ are pre-trained linear projection matrices for query, key, and value, respectively, and are maintained frozen all along the editing process. $W_q^t$ is the linear projection to be fine-tuned for the target branch. And $x$ is the hidden states, the output of the previous layer.

Note that distinct from customization methods\cite{ruiz2023dreambooth, gal2022image, kumari2023multi, cao2023concept}, that optimize the U-Net neurons for the preservation of identity, we tune these parameters to allow for more fine-grained image modification control (e.g. Fig. \ref{fig:pca}).

\subsubsection{CLIP Loss.} Given that the CLIP model builds the bridge between language and vision modality, we devise the supervisor following a straightforward intuition that good edit results should be close to the target prompt in CLIP embedding space. Due to that the CLIP vision model can only process the image in RGB space, we apply CLIP loss at the end of the pipeline as illustrated by Fig. \ref{fig:dual_branch_pipeline}. We directly set the simple text-to-image cosine similarity as our loss $\mathcal{L}_{clip}$, formulated as:

\begin{equation}
\begin{aligned}
\mathcal{L}_{clip} = 1 - \frac{{I_t \cdot T}}{{\|I_t\| \|T\|}}
\end{aligned}
\end{equation}
where $I_t$ and $T$ are CLIP-encoded embeddings of the edited image and text prompt.

On the other hand, we design a regularization loss to regularize the modification guided by CLIP loss. We set the loss between the last sampled latent of two branches, $z_0^t$ and $z_0^s$ as follows:
\begin{equation}
\begin{aligned}
\mathcal{L}_{reg} = \|z_0^t - z_0^s \|^2
\end{aligned}
\end{equation}

And our final loss becomes:
\begin{equation}
\begin{aligned}
\label{eq: clip_loss}
\mathcal{L} = \mathcal{L}_{clip} + \lambda\mathcal{L}_{reg}
\end{aligned}
\end{equation}
where $\lambda$ is defined as reconstruction strength to constrain excessive modification to prevent edit results from collapsing. We empirically set the $\lambda$ in the range of 0 to 10.0 depending on tasks and the change to be applied. Note that, although the reconstruction loss and our feature-sharing pipeline both contribute to the preservation of the original image, they play different roles. Our ablations show that the shared feature lays the foundation of the overall texture and appearance while reconstruction loss strays the turbulence of changes. More details are provided in Sec. \ref{subsubsec: Regularization Loss v.s. Feature Sharing}.

\subsubsection{Random-gateway Optimization Mechanism.} Although our insight is straightforward that using the CLIP model to supervise the editing process, it is rather burdensome to back-propagate the gradient signal through the whole diffusion process with the naive algorithm. Consider a single step $t$ in the sampling process (Eqn. \ref{eq:sample} and Fig. \ref{fig:random-gateway} (a)), the noised latent code $z_t$ is fed as input into the U-Net, together with time embedding $t$ and text embedding $C$ to compute the noise $\epsilon$. The denoised latent is a linear summation of two parts, the predecessor latent and the predicted noise. It is obvious that when we apply CLIP loss at the end of the sampling process, the U-Net will iteratively forward by $T$ times which can lead to a heavy memory burden during back-propagation for we need to deal with all these stacked U-Net gradients. 

To tackle the problem, we propose an efficient optimization method named \textit{Random-Gateway} to alleviate memory usage while maintaining the effectiveness of CLIP guidance. Specifically, we detach the gradients flow into the \textit{noise term $\epsilon_\theta$} at most steps Fig. \ref{fig:random-gateway}(b) and only leave several steps as gradient \textit{\textbf{gateways}} Fig. \ref{fig:random-gateway}(a). Formally, we have each sampling step as:

\begin{equation}
\begin{aligned}
\text{$samp_t$}:= 
\begin{cases}
  k_1z_t + k_2\epsilon_\theta, & \text{if } t \in t_{gateway}, \\
  k_1z_t + \text{sg}(k_2\epsilon_\theta), & \text{otherwise}.
\end{cases} 
\end{aligned}
\end{equation}
where $k_1$, $k_2$ denote the two $\alpha$-related coefficients of the two add terms in Eqn. \ref{eq:sample}, sg($\cdot$) denotes the stop gradient operator ($detach$ in Pytorch).

\begin{figure}[t!]
\centering
\includegraphics[width=1\textwidth]{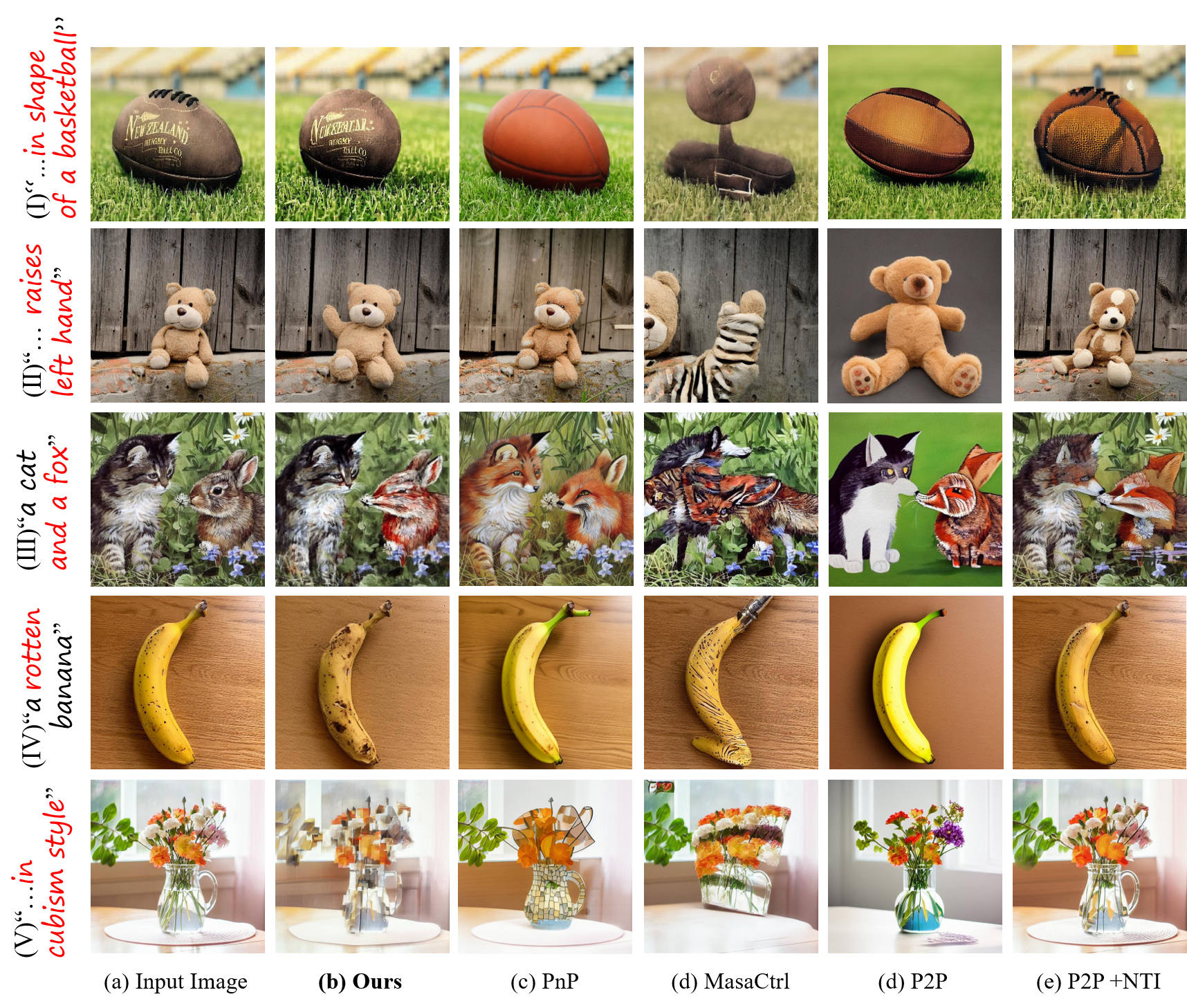}
\caption{\textbf{Qualitative comparisons.} E4C outperforms other leading methods in visual outputs across all five task settings.}
\label{fig:qualitative_comparison}
\vspace{-4mm}
\end{figure}

At those steps that are stopped gradients, U-Net parameters will not be updated immediately but the supervision signal (the gradient flow) can still be passed to the previous step through the \textit{latent term $z_t$}. To this end, we only back-propagate U-Net for times equivalent to the number of those gateways, thus constraining the memory usage to a controllable level. Although the algorithm inevitably weakens the gradient flow in quantification due to the other term being eliminated, we have experimentally demonstrated that it is enough for delicate content editing and achieving a strong consistency with text prompts. In practice, we uniformly and randomly select five $t_{gateway}$s without repetition for every loop. We experimentally demonstrate that random strategy suits more situations than deliberate selection. See the Appendix for More information about the algorithm.

Note that as we are using the shared U-Net across all timesteps, updating the U-Net at every single gateway still receives the bidirectional optimizing signals. In this context, we actually decompose the optimization for the entire editing process into fitting the U-Net to several denoising steps. It also explains the sensibility of our method.

To summarize, we demonstrate our efficiency in utilizing CLIP guidance from 3 perspectives: (\romannumeral1) We only optimize a small subset of parameters (the self-attention features) to accommodate delicate manipulations. (\romannumeral2) We improve the editability and text alignment in a thrifty way that only caters to several timesteps. (\romannumeral3) We significantly reduce memory usage compared to other CLIP-guided methods\cite{kim2022diffusionclip, samuel2023generating}, making it possible to use consumer-grade GPU cards with 24GiB memory. Additional experimental statistics are provided in the Appendix.

\begin{table}[t!]
    \centering
    \caption{\textbf{Quantitative comparisons with other methods.} Note that PTP+NTI\cite{hertz2022prompt, mokady2023null} gets higher DINO scores than ours for the observation that it is prone to over-protect the source image, especially in style transfer and non-rigid tasks whereas ours tends to perform the edits.}
    \begin{tabular}{ccccc}
      \toprule
      \multirow{2}{*}{Method} & \multicolumn{2}{c}{Structure-Preserved} & \multicolumn{2}{c}{Non-rigid} \\
      \cmidrule(r){2-3} \cmidrule(r){4-5}
       & CLIP Score(\textuparrow) & DINOv2 similarity(\textdownarrow) & CLIP Score(\textuparrow) & DINOv2 similarity(\textdownarrow) \\
      \midrule
      P2P + NTI  & \underline{28.33} & \textbf{0.026} & 24.31  & \textbf{0.025}\\
      P2P        & 23.86 & 0.157 & 22.65  & 0.132 \\
      PnP        & 27.83 & \underline{0.083} & 20.97  & 0.101 \\
      MasaCtrl   & 24.41 & 0.112 & \underline{25.35}  & 0.093 \\
      Ours       & \textbf{34.55} & 0.084 & \textbf{30.57}  & \underline{0.074} \\
      \bottomrule
    \end{tabular}
    \label{tab:my_label}
    \vspace{-4mm}
\end{table}

\section{Results}

\subsection{Implementation details}
We apply the proposed method to the most widely used pre-trained model Stable Diffusion model with official checkpoints v1.5. We empirically set the feature-sharing scheme for the last 6 self-attention layers and timesteps from 5 to 50 in a 50-step case. For different editing requirements, we set $\lambda$ in the range of $[1.0, 10.0]$ depending on the edits' spatial scope and extent. For the CLIP guidance, we set the number of random gateways as 5 for a 24GiB memory device, and the sampling process is empirically looped 1 to 5 times depending on cases. All experiments are conducted on a single NVIDIA 3090 GPU. More details are provided in the Appendix.

\subsection{Comparisons with previous methods}

\subsubsection{Benchmark dataset and baselines}
\label{subsec:experiment setup}
Most existing text-based image editing datasets lack categorization into editing tasks and do not adequately evaluate the methods' comprehensive ability or potential application scenes. Based on the textual editing benchmark TEdBench proposed by Imagic\cite{kawar2023imagic} and \textit{Wild-TI2I} by PnP\cite{tumanyan2023plug} and additional web-crawled images, we re-phrase the edit instructions to extend the editing scope and varieties and re-organize the image-prompt pairs into 5 types of editing tasks including object replacement, object attribute manipulation, style transfer, creature pose change, item shape change (20 image-prompt for each and 100 in total). We compare E4C with competing methods Plug-and-play\cite{tumanyan2023plug}, Prompt-to-prompt\cite{hertz2022prompt} (w/ and w/o Null-Text inversion\cite{mokady2023null}), and MasaCtrl\cite{cao2023masactrl}, all methods are based on the foundation model SD1.5.

\begin{figure}[t!]
\centering
\includegraphics[width=1\textwidth]{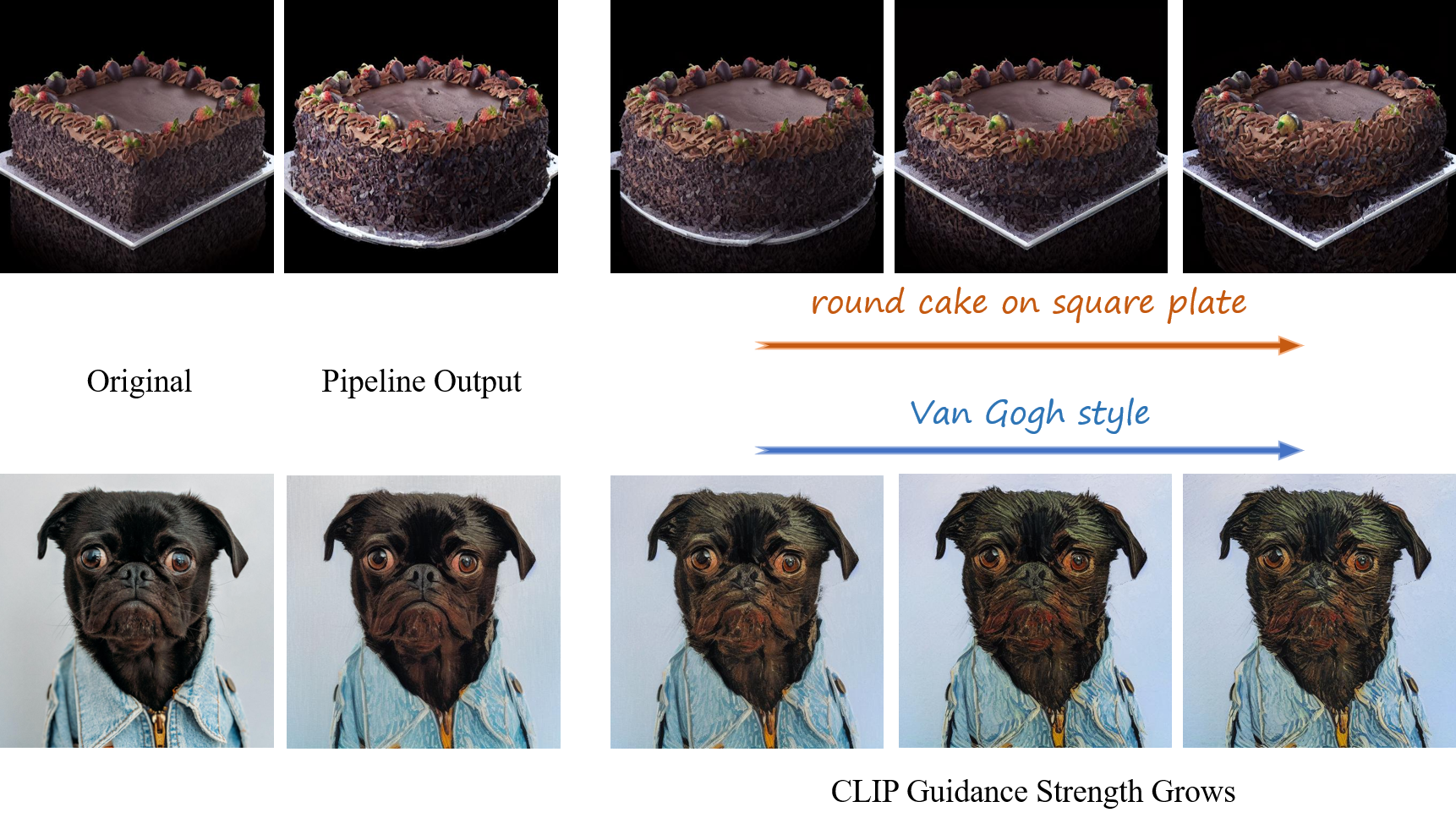}
\caption{\textbf{Edit results as CLIP guidance strength grows.} From left to right displays the original image, direct editing result, and as sampling process loops 1, 2, and 3 times. As CLIP strength grows, modification is steadily applied for better text alignment and more reasonable representations.}
\label{fig:ablation_clip}
\vspace{-3mm}
\end{figure}

\subsubsection{Qualitative evaluation}
As Fig. \ref{fig:qualitative_comparison} shows, our methods get super competitiveness across all five types of edit tasks versus other competing methods even in their strength domain. Due to the imperfect reconstruction of naive DDIM inversion\cite{song2020denoising}, P2P\cite{hertz2022prompt} without the support of NTI\cite{mokady2023null} fails in mapping the authentic colors and structure, let alone the alignment with the target prompt. PnP\cite{tumanyan2023plug} and P2P-NTI both perform relatively well in preserving the overall structure and texture information, yet they can barely deal with shape change and pose change. For tasks of shape change and pose change, only ours and MasaCtrl\cite{cao2023masactrl} can alter the structure of the object but the latter fails to generate the right segmentation shape of "basketball"(the row \RomanNumeral1) or correctly apply the pose change to the left part of teddy bear (the row \RomanNumeral2). In a multi-object scenario (the row \RomanNumeral3), though PnP and P2P-NTI can generate identifiable fox icons, they cannot preserve the cat identity unchanged (fur color and mouth shape affected by "fox"), while our methods automatically locate the area changes should be applied with no affiliate conditions like masks or segmentation map. For attribute and style manipulation (the rows \RomanNumeral4 and \RomanNumeral5), only our method can correctly control the feature of the image with adjective words "rotten" and "cubism style" (PnP generates cubes on vase but still poses a difference with the art style "cubism"). More qualitative results are included in the Appendix.

\vspace{-3mm}

\subsubsection{Quantitative evaluation}
We conduct quantitative analysis on our method and baseline models with the CLIP Score\cite{radford2021learning} and DINO-ViT self-similarity distance\cite{tumanyan2022splicing} to evaluate the performance on aligning to text prompt and preserving source image information, respectively. As mentioned in Sec. \ref{subsec:experiment setup}, we compute metrics on all our five tasks and average the object replacement, object attribute manipulation, style transfer as structure-preserved task and creature pose change, and item shape change as non-rigid editing tasks. Table. \ref{tab:my_label} illustrates that our method achieves much higher CLIP scores than the others while maintaining relatively satisfying fidelity. We can observe that P2P\cite{hertz2022prompt} (w/ and w/o NTI\cite{mokady2023null}) and PnP\cite{tumanyan2023plug} perform better on structure-preserved tasks while MasaCtrl\cite{cao2023masactrl}, in contrast, behaves strength on pose or shape change. Note that P2P+NTI surpasses us in image similarity. Based on our visualization results, we hypothesize that P2P+NTI is more inclined to preserve the original image and less likely to generate new texture styles compared to our method in style transfer and non-rigid tasks.

\begin{figure}[t!]
\centering
\includegraphics[width=1\textwidth]{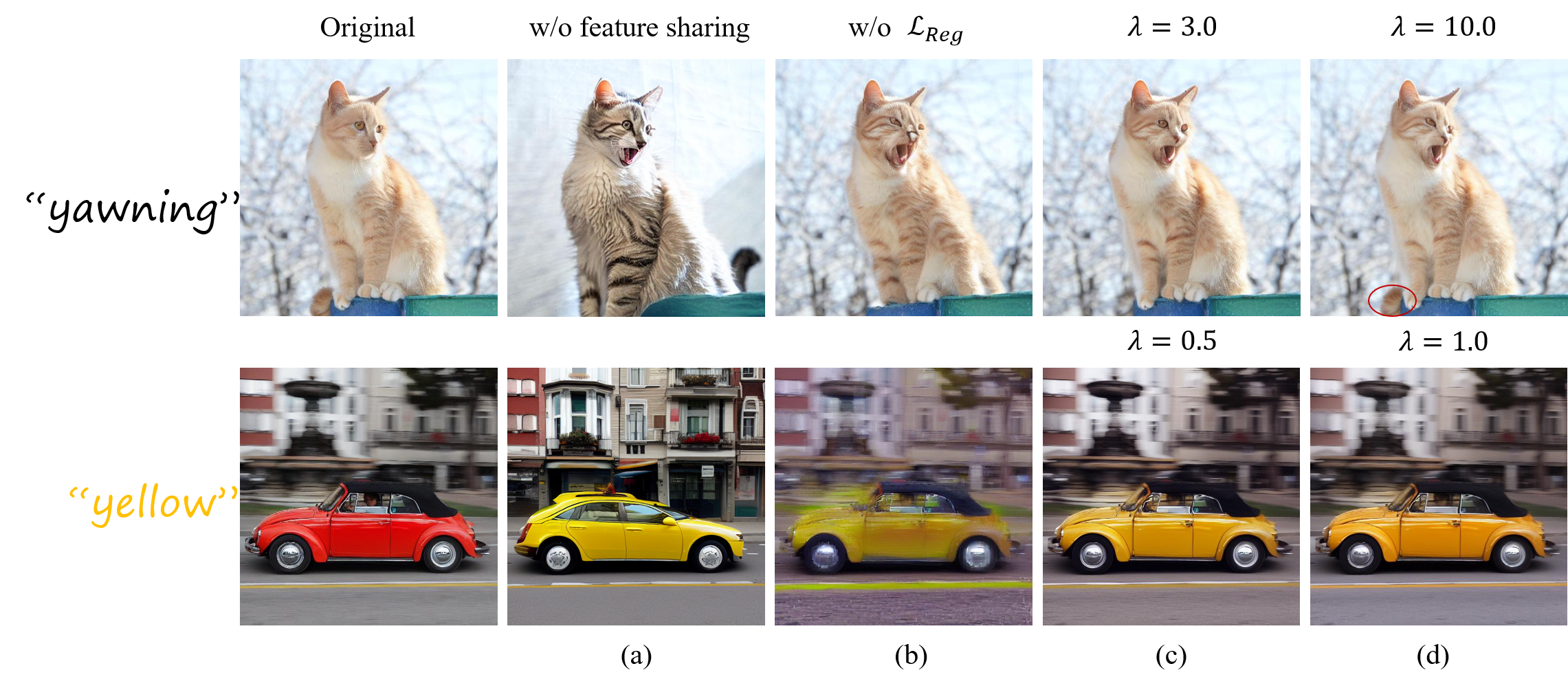}
\caption{\textbf{Orthogonal effectiveness of feature-sharing scheme and $\mathcal{L}_{reg}$.} We illustrate that the two components play different roles in preserving fidelity and showcase the influence of different reconstruction strengths.}
\label{fig:ablation_loss_reg}
\vspace{-4mm}
\end{figure}

\subsection{Ablation Study}
\label{ablation}
\subsubsection{Impact of CLIP Guidance.}
As noted above, the clip guidance is applied at the end of our pipeline and as the sampling process loops the guidance strength should be moderately enhanced, bringing the edit results closer to the text prompt semantically. Fig. \ref{fig:qualitative_comparison} shows the comparison between the resulting output by editing branch and those under CLIP guidance for 1, 2, and 3 loops. The top row illustrates that the direct output successfully turns the shape of the cake into a round though, it fails to keep the plate square and the reflection of the cake disappears. With CLIP guidance growing, it automatically locates the modification on the cake solely and fixes the plate and reflection. The same is true for style transferring (bottom row), only the texture of the painting turns out from the realism style, and under the supervision of the CLIP model, it appears the symbolic curves and vibrant colors of van Gogh's works.

\subsubsection{Regularization Loss v.s. Feature Sharing Scheme.}
\label{subsubsec: Regularization Loss v.s. Feature Sharing}
Both the feature-sharing scheme and $\mathcal{L}_{reg}$ are strategies for proving the fidelity to the original, and we demonstrate that they perform orthogonal effectiveness. As is shown in Fig. \ref{fig:ablation_loss_reg} without feature sharing (column a), the texture of the cat's fur and the shape of the car is dramatically changed, as well as their background. On the other side, CLIP guidance may put too much influence (column b) on the pose (the twisted cat's expression and ambiguous nose) and the color (the center of the car is blurred), in the absence of $\mathcal{L}_{reg}$. Also, we empirically set different reconstruction strengths according to the situation. In the context of pose change in a small spatial area like turning a cat from mouth closed to yawning, setting $\lambda$ to 10.0 can retain the micro details like the cat's tail (red circled in column d cat).

% \subsection{Limitations and Extensions}
% Although our editing method has shown effectiveness in various types of editing tasks, we have identified some limitations that require further improvements. While our feature-sharing pipeline performs well in common scenarios, we have observed its weakness in the human face domain, especially when editing high-resolution and detailed face images. We hypothesize that this is due to the incompatible expressiveness of low-resolution features of U-Net, resulting in the loss of high-density information from the source image during the editing process. Moreover, our methods can be seamlessly applied to more powerful pre-trained models like SDXL, we provide the edit results of the SDXL version in the Appendix.

\section{Conclusions}
In this work, we propose \textbf{E4C}, catering to multiple types of complex tasks including both structure-preserved and non-rigid editing. In detail, we build a dual-branch pipeline that allows for adaptive SA feature sharing to selectively preserve the structure or texture information. Benefiting from our random-gate optimization mechanism, we can further leverage the CLIP guidance to significantly enhance the editability and alignment to the target prompt. With no affiliate conditions (masks, segmentation maps, etc.), our method shows inherent strength in handling hard samples, for example, spontaneously localizing the editing part in multi-object scenarios and manipulating poses and shapes of items. We believe that our work pushes the boundaries and presents more potential for image editing, especially under pure text guidance. We hope it will inspire and motivate future researchers to develop more general and concise editing pipelines.

% ---- Bibliography ----
%
% BibTeX users should specify bibliography style 'splncs04'.
% References will then be sorted and formatted in the correct style.
%
\bibliographystyle{splncs04}
\bibliography{main}

\clearpage

\appendix

\section{Summary}
\label{sec:summary}
In this appendix, we present more details about the background of the diffusion model, the detailed algorithm and efficiency comparison of the \textit{random-gateway} mechanism mentioned in Sec.3.3 of the main paper, implementation details of experiments, and more qualitative visual results. This appendix is composed as follows:
\begin{itemize}
    \item In Sec. \ref{sec:preliminaries}, we present the essential preliminaries about the diffusion model.
    \item In Sec. \ref{sec:efficient clip guidance}, we illustrate the fake codes of the random-gateway mechanism, the comparison with other diffusion-clip combined methods on memory and time-consuming, and the ablation of the \textit{random} strategy.
    \item In Sec. \ref{sec:experiments}, we illustrate more implementation details as we conduct the experiments and comparison with other methods.
    \item In Sec. \ref{sec:qualitative}, we demonstrate more visual results of our methods in various tasks.
    \item Finally, we discuss the limitations of our methods in Sec. \ref{sec:limitations}.
\end{itemize}

\section{Background and Preliminaries of the Diffusion model}
\label{sec:preliminaries}
Diffusion models\cite{ho2020denoising, song2020denoising} are generative models that transform image generation into a denoising process. The diffusion process is composed of two main processes, namely the forward and reverse which add and remove Gaussian noise from the data sample, respectively. The forward process can be formulated as: 

\begin{align}
    q(x_t|x_0) = \mathcal{N}(x_t; \sqrt{\overline{\alpha}}, (1 - \overline{\alpha})_t\textit{I}
\end{align}

where $\alpha$ is the noise scheduler that controls the noise strength and through such a formula we can get the $x_T$ which is close to the Gaussian sample $\mathcal{N}(0, 1)$.

The reverse process starts from $x_T$ and follows an iteration denoising equation:

\begin{align}
    p_\theta(x_{t-1}|x_t) = \mathcal{N}(x_{t-1}; \mu_\theta(x_t, t); \sigma_t) \\
    \mu_\theta(x_t, t) = \frac{1}{\sqrt{\alpha_t}}(x_t - \epsilon\frac{1 - \alpha_t}{1 - \overline{\alpha_t}})
    \label{eq:sample}
\end{align}

where $\sigma$ is a constant that relates to the time-step $t$ and $\epsilon$ is the noise term. In the training procedure, we preset the $\epsilon$ and use a neural network (U-Net\cite{ronneberger2015u})$\epsilon_\theta$ to predict the noise, and we train the network to minimize the difference between the estimated noise $\epsilon_\theta(x_t, t)$ and $\epsilon$ by a mean-squared error:
\begin{align}
    \mathcal{L} = E(\| \epsilon - \epsilon_\theta(x_t, t)\|)
\end{align}

Note that the noise estimator $\epsilon_\theta$ can receive other conditions like text-prompt embeddings (Stable Diffusion\cite{rombach2022high}) and the predicted noise can be extended to $\epsilon_\theta(x_t, t, c)$. Following Eqn. \ref{eq:sample} we can finally get the denoised $x_0$ which are images that comply with the training samples and additional conditions.

\section{Details of Efficient CLIP Guidance}
\label{sec:efficient clip guidance}

% \begin{algorithm}
% \begin{algorithmic}
% \caption{Random-gateway optimization}
% \KwData{$P_t$: the target prompt, $\epsilon_\theta$: pre-trained U-Net, \\
%         $z_T$: result of DDIM inversion, $n_{steps}$: number of DDIM scheduler timesteps, \\
%         $N$: number of training loops, $n$: number of gateway in a single loop, \\
%         $L$: CLIP loss, $optimizer$: optimizer for updating parameters}
% \KwResult{$z_0$}
% \For{$i=N...1$}{
%   \State $t_{\text{gateway}} \gets \text{np.random.choice}(\{ \text{np.arange}(n_{\text{steps}}), n \})$ \\
%   \For{$t=T...1$}{
%     $\epsilon_t = \epsilon_\theta(z_T, t, P_t)$ \\
%     $z_{t-1}=k_1z_t + k_2\epsilon_t$ \\
%     \If{$t\notin t_{\text{gateway}}$}
%       {$\epsilon_t = \text{sg}(\epsilon_t)$}
%     \EndIf
%   }
%   $loss = L(z_0, P_t)$ \\
%   $loss.backward()$ \\
%   $optimizer.step()$
% }
% \end{algorithmic}
% \end{algorithm}

\begin{algorithm}
\caption{Random-gateway optimization}
\begin{algorithmic}[1]
\STATE \textbf{Input:}
\STATE $P_t$: the target prompt
\STATE $\epsilon_\theta$: pre-trained U-Net
\STATE $z_T$: result of DDIM inversion
\STATE $n_{\text{steps}}$: number of DDIM scheduler timesteps
\STATE $N$: number of training loops
\STATE $n$: number of gateway in a single loop
\STATE $L$: CLIP loss
\STATE $optimizer$: optimizer for updating parameters
\STATE \textbf{Output: $z_0$}
\FOR{$i=N$ \textbf{to} $1$}
  \STATE $t_{\text{gateway}} \gets \text{np.random.choice}(\{ \text{np.arange}(n_{\text{steps}}), n \})$
  \FOR{$t=T$ \textbf{to} $1$}
    \STATE $\epsilon_t = \epsilon_\theta(z_T, t, P_t)$
    \STATE $z_{t-1}=k_1z_t + k_2\epsilon_t$
    \IF{$t\notin t_{\text{gateway}}$}
      \STATE $\epsilon_t = \text{sg}(\epsilon_t)$
    \ENDIF
  \ENDFOR
  \STATE $loss = L(z_0, P_t)$
  \STATE $loss.\text{backward}()$
  \STATE $optimizer.\text{step}()$
\ENDFOR
\end{algorithmic}
\end{algorithm}

\subsubsection{Gateway Selection Strategy.} For each training loop, we need to choose timesteps as gateways, and we conduct an ablation study on how to choose the appropriate steps. In the context of a 50-timesteps sampling process and 5 gateways for each loop, we pre-set four selection strategies, randomly choosing from the former or latter 25 steps, selecting one from the 10-step intervals (from $[0, 10)$, $[10, 20)$, ..., $[40, 50)$ choosing one for each) and random. We demonstrate that random strategy can cater to most situations both qualitatively and quantitatively.
As the Table. \ref{tab:ablation} shows, that choosing from the former of the latter half timesteps can cause inferior results as there may be bias in the different denoising stages and get a lower CLIP score (though they get a little higher in DINO similarity, we care more about the ability to apply correct editing content). Choosing from 10-timestep intervals and a random strategy have similar performances for they have similar sample distribution in most observation cases. And in practice, both can generate relatively good edit results. Note that the "random" means we do not need to deliberately choose our gateways.

\subsubsection{Demonstration on \textit{Efficiency}.} We claim that our method demonstrates efficiency compared to the other CLIP-guided diffusion models on both memory and time-consuming. We conduct experiments on two other CLIP-integrated diffusion models, DiffusionCLIP and SeedSelect. DiffusionCLIP and ours share the objective that utilizing the CLIP model to improve the text alignment of the edit results. On the other hand, SeedSelect aims at generating rare subjects and hard samples with higher productivity (like the infamous hands' collapse). Note that ours and SeedSelect use the same pre-trained latent diffusion and CLIP models and DiffusionCLIP uses the domain-specific diffusion models.

Table. \ref{tab:resources} shows that our method outperforms SeedSelect\cite{samuel2023generating} and DiffusionCLIP\cite{kim2022diffusionclip} in both memory usage and Time Spending. We compute the memory and time consumption for one sampling loop and the model updating process (excluding the inversion process). Specifically, DiffusionCLIP is built on a group of diffusion models for a few domains (e.g., human face, church, bedroom, dog face, etc.). We conduct experiments using the pre-trained checkpoints provided in their released codes. We analyze resources requirements on editing 256x256 resolution images (row 1 in Table. \ref{tab:resources}) and as claimed in their paper, the 512x512 editing process should be consuming twice the memory usage and 4 times the time consumption. For Select, we test with the officially recommended at-least 20 timesteps and it takes about 35 GiB, and for more timesteps, it could take more than 50GiB and more optimization time. For all testing methods, we use gradient checkpointing\footnote{https://pytorch.org/docs/stable/checkpoint.html}, and for ours, we specifically use the xformers attention implementations\footnote{https://github.com/facebookresearch/xformers}.

\section{Details of Experiments}
\label{sec:experiments}
\subsection{Dataset Preparation.} 
As mentioned in the main paper, we collected 40 images from TEDBench\cite{kawar2023imagic} and Wild-TI2I\cite{tumanyan2023plug} and crawled another 20 from the Internet. The main content varies from the items to living creatures in styles including photo-realistic and drawings. We first utilize BLIP\footnote{https://huggingface.co/Salesforce/blip-image-captioning-large} to caption the images to get the source description of the images and rephrase them to fit our editing requirements. Then, according to task types(object replacement, object attribute manipulation, style transfer, pose changes, and shape changes), we alter or add certain words to the original. For example, replacing the word "rose" with "sunflower" in "a dog holding a rose in the mouth" is an object replacement task. Note that the object replacement task aims to swap the items or creatures with another in similar shape or types (e.g. different sorts of flowers, birds), while the attribute emphasizes changing certain parts or statuses of the object like the color. In some cases, the difference between two tasks can be blurred like changing the liquid in a bottle that could be either regarded as the attribute of the bottle of sth. or the replacement of the liquid. Moreover, pose changes could be "sitting" to "running", "standing" to "jumping" or more delicate manipulation like if the mouth closed or opened. As for shape-changing, besides the explicit adjectives like "round", and "triangle" ..., we use "in the shape of sth." to be an indirect hint. Our final dataset is composed of 100 $(image, P_s, P_t)$ triplets, 20 for each task.

\subsection{Experiment Settings}
The source images could be in different resolutions: most of them are 512x512 (the standard shape of SD output) the smallest is 256x256 and the largest is 1250x1250. For preprocessing, we used the interpolate function in Pytorch\footnote{https://pytorch.org/docs/stable/generated/torch.nn.functional.interpolate.html} and normalized them to range of $[-1, 1]$. 

\begin{table}[t!]
  \centering
  \caption{\textbf{Different hyer-parameters setting for varying tasks.} Note the hyper-parameters for experiment settings are not absolute, they may depend on the degree and spatial area that the changes applied.}
  \begin{tabular}{ccc}
    \toprule
    Tasks & learning rate & $\lambda$ \\
    \midrule
    object attribute manipulation & 1e-4 & 1.0 \\
    object replacement & 5e-4 & 0.5 \\
    style transfer & 1e-4 & 0.0 to 1.0 \\
    pose change & 1e-3 & 1.0 to 10.0 \\
    shape change & 5e-4 to 1e-3 & 0.5 to 1.0 \\
    \bottomrule
    \label{tab:hyper-params}
  \end{tabular}
\end{table}

\subsubsection{Ours.} We DDIM inversion over the timesteps $T=50$, classifier-free guidance 7.5 and we use the null prompt " as the input text. The setting is maintained for the reconstruction process. For CLIP optimization, we use the CLIP ViT-H-14 model which is released at\footnote{https://huggingface.co/laion/CLIP-ViT-H-14-laion2B-s32B-b79K/tree/main}. We set projection matrices ($W_k$ and $W_v$ for structure-preserved and $W_q$ for non-rigid tasks) in all self-attention modules of U-Net as our optimizing parameters. The Adam optimizer ($betas=(0.9, 0.999)$) is applied for gradients updating. For catering to varying edit tasks, we empirically set different learning rates and reconstruction strengths ($\lambda$) as illustrated in Table. \ref{tab:hyper-params}.

\subsubsection{Comparison Methods.} For comparison, we follow the released official implementations of Prompt-to-prompt\footnote{https://github.com/google/prompt-to-prompt}, NTI\footnote{https://github.com/google/prompt-to-prompt}, Plug-and-Play\footnote{https://github.com/MichalGeyer/plug-and-play} and MasaCtrl\footnote{https://github.com/TencentARC/MasaCtrl}. Specifically, P2P provides three kinds of attention controllers for replacement, refinement, and reweighting. We use AttentionReplace for object replacement and AttentionRefine for object attribute manipulation while applying AttentionReweight on specific editing phrases like the object word, the adjective, "in xx styles", and "in the shape of xx". Also, the local blending is applied to the words of the object to edit. 

\begin{table}[t!]
  \centering
  \caption{\textbf{Comparisons of Resources consuming.} We present the sizes of the foundation model and optimizing parameters for each method and their time and memory consumption. Note the time is computed for one forward sampling process and the optimizer updating parameters.}
  \begin{tabularx}{\textwidth}{lXXXX}
    \toprule
    Method & Model (GiB)& Parameters (GiB) & Memory (GiB) & Time (second) \\
    \midrule
    DiffusionCLIP (256x256) & 0.167 & 0.167 & 23  & 42 \\
    DiffusionCLIP (512x512) & 2.1  & 2.1 & 46 & 166 \\
    SeedSelect (512x512) & 3.3 & 3.3 & 35 at least & 60 \\
    Ours (512x512) & 3.3 & \textbf{0.02} & \textbf{17} & \textbf{17} \\
    \bottomrule
    \label{tab:resources}
  \end{tabularx}
\end{table}

\begin{table}[t!]
    \centering
    \caption{\textbf{Ablation Study on Gateway Selection Strategy.} Qualitative comparisons on different gateway selection strategies. Note that all results are averaged on 10 different random seeds.}
    \begin{tabular}{ccccc}
      \toprule
      \multirow{2}{*}{Method} & \multicolumn{2}{c}{Structure-Preserved} & \multicolumn{2}{c}{Non-rigid} \\
      \cmidrule(r){2-3} \cmidrule(r){4-5}
       & CLIP Score(\textuparrow) & DINOv2 similarity(\textdownarrow) & CLIP Score(\textuparrow) & DINOv2 similarity(\textdownarrow) \\
      \midrule
      former 25 steps  & 31.65 & \textbf{0.066} & 26.97  & \underline{0.054}\\
      latter 25 steps        & 31.47 & 0.071 & 25.56  & \textbf{0.051} \\
    10-step intervals & \underline{32.98} & \underline{0.069} & \underline{29.07}  & 0.068 \\
      Random       & \textbf{33.10} & 0.088 & \textbf{29.25}  & 0.070 \\
      \bottomrule
    \end{tabular}
    \label{tab:ablation}
    \vspace{-4mm}
\end{table}

\section{More Qualitative Results}
\label{sec:qualitative}
\begin{figure}[t!]
\centering
\includegraphics[width=1\textwidth]{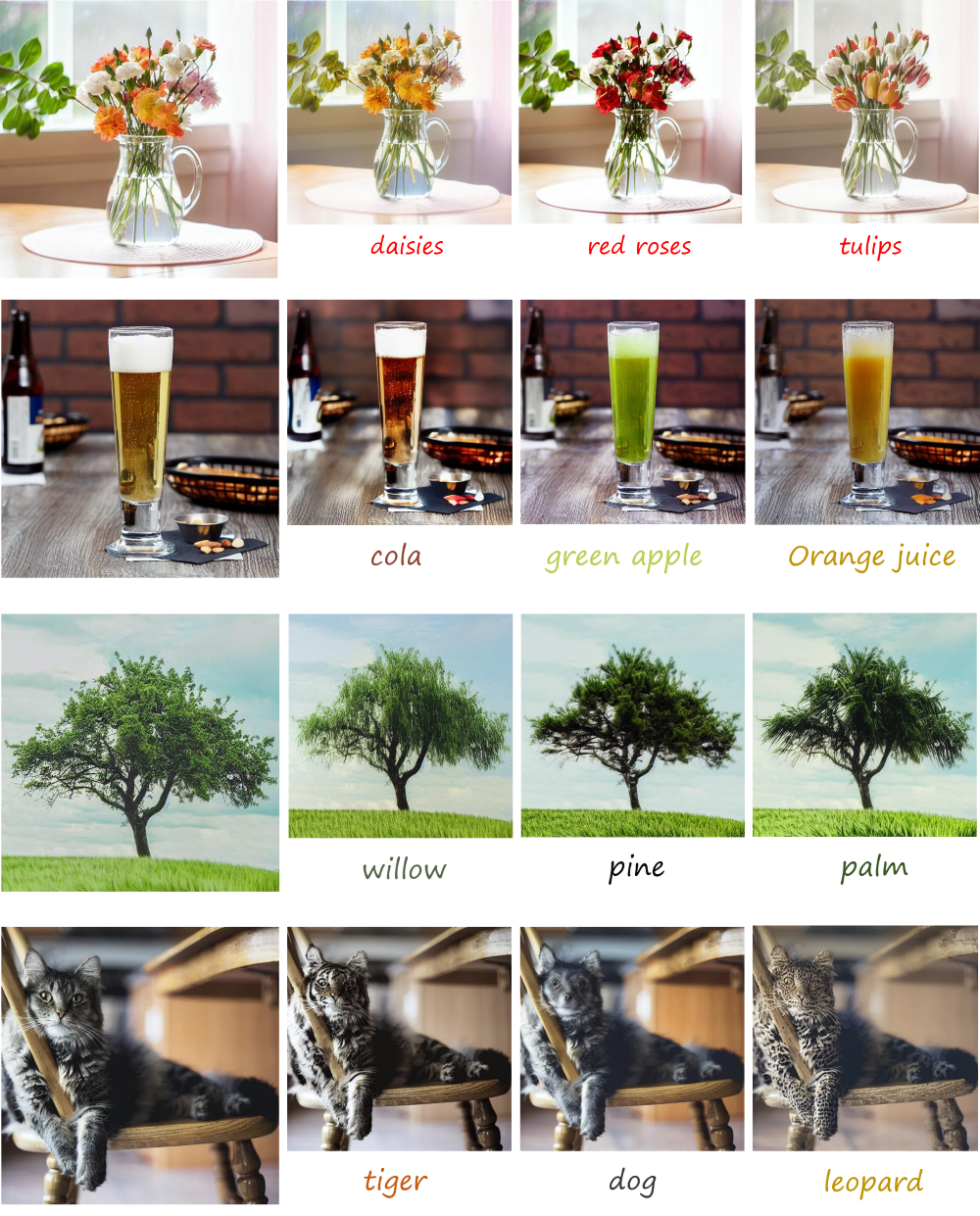}
\caption{\textbf{Object replacement.} Our method can replace the type of the object with another that shares the same pose and layout.}
\label{fig:object_replacement}
\end{figure}

\begin{figure}[t!]
\centering
\includegraphics[width=1\textwidth]{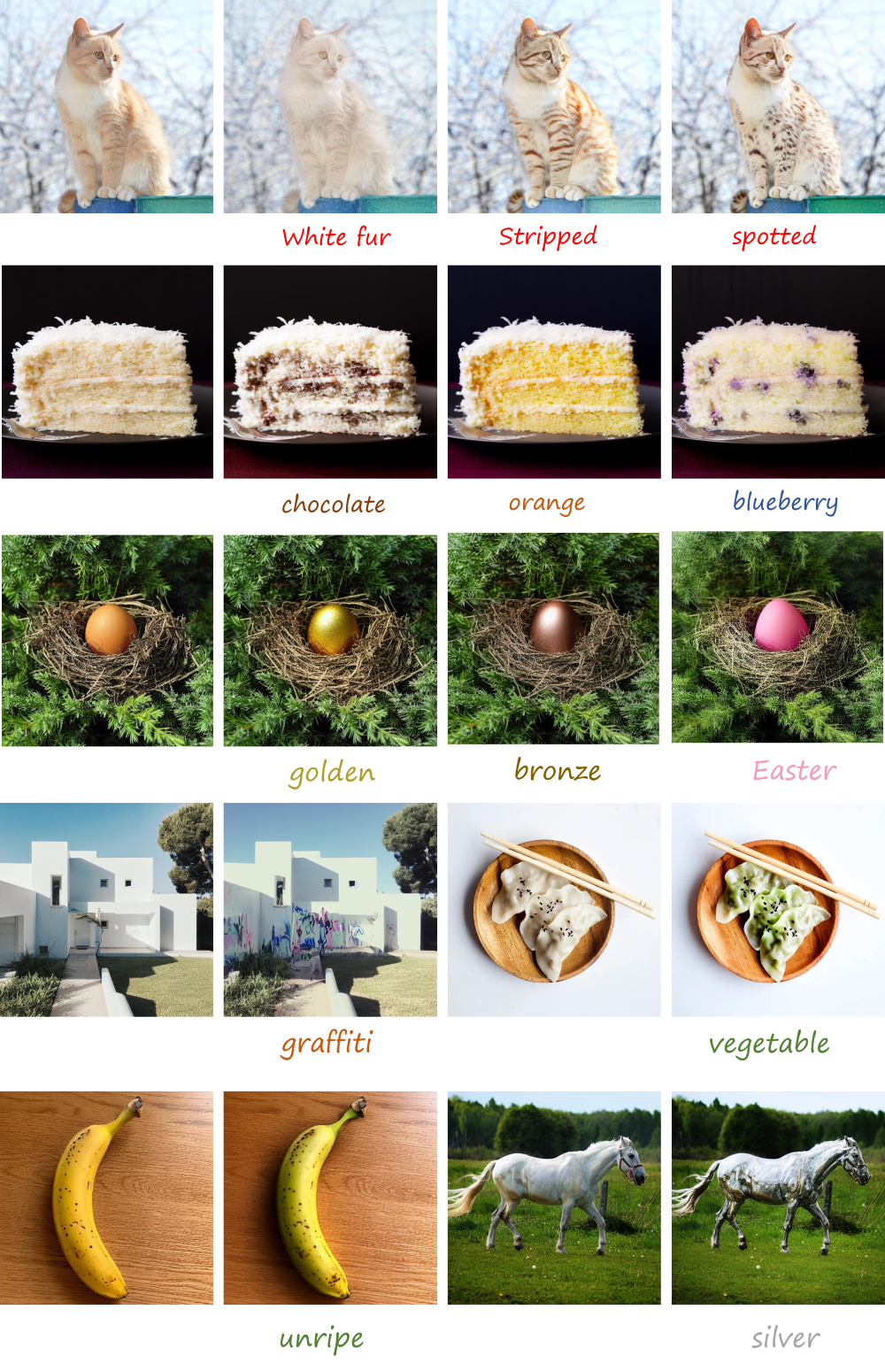}
\caption{\textbf{Object attribute manipulation.} Our method can change the texture and appearance of the object and its status.}
\label{fig:attribute_change}
\end{figure}

\begin{figure}[t!]
\centering
\includegraphics[width=1\textwidth]{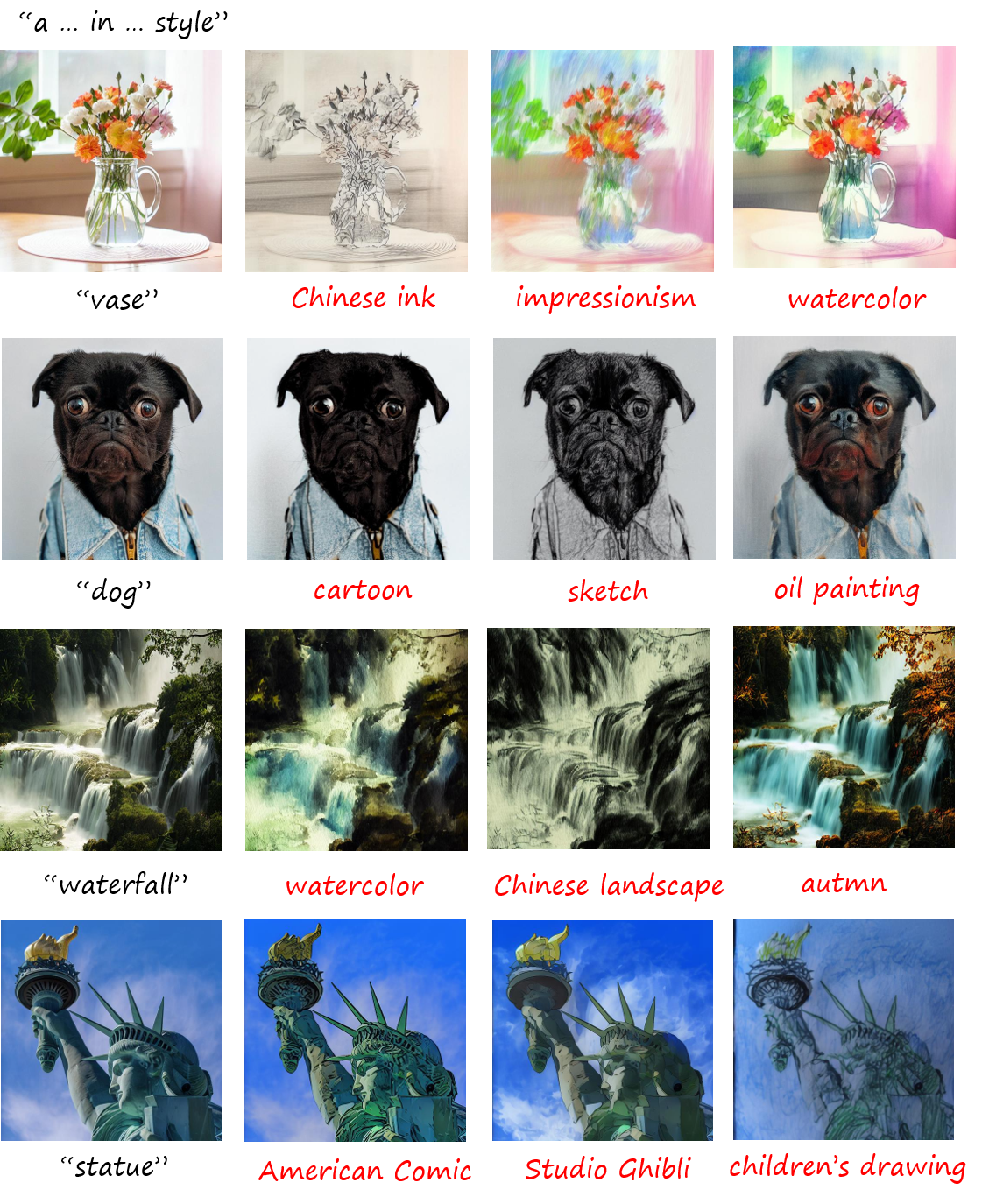}
\caption{\textbf{Text-driven style transfer.}}
\label{fig:style_transfer}
\end{figure}

\begin{figure}[t!]
\centering
\includegraphics[width=1\textwidth]{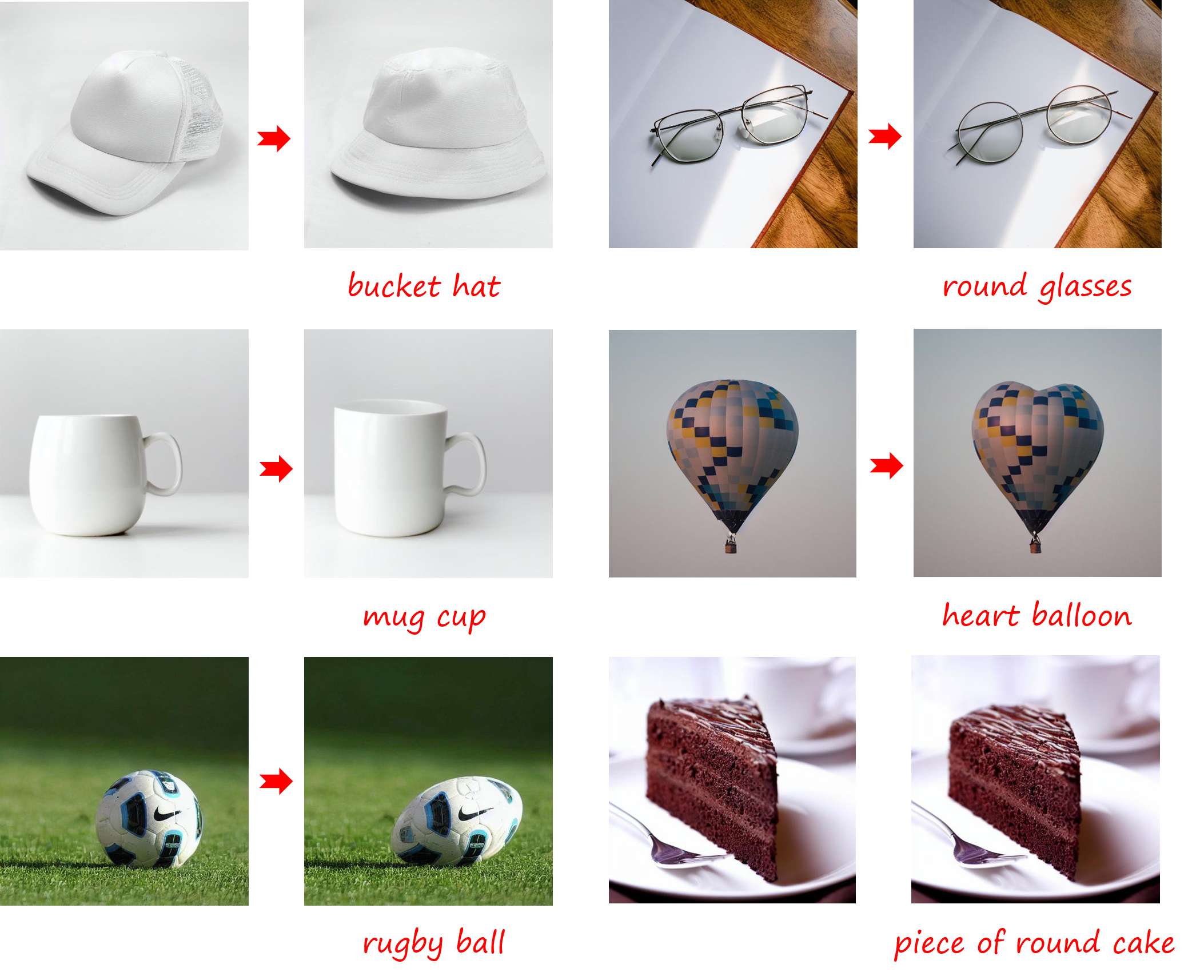}
\caption{\textbf{Items shape change.} Our method can perform edits on object shapes. This feature can be applied to generate type variations of hat, cup, etc.}
\label{fig:shape_change}
\end{figure}

\begin{figure}[t!]
\centering
\includegraphics[width=1\textwidth]{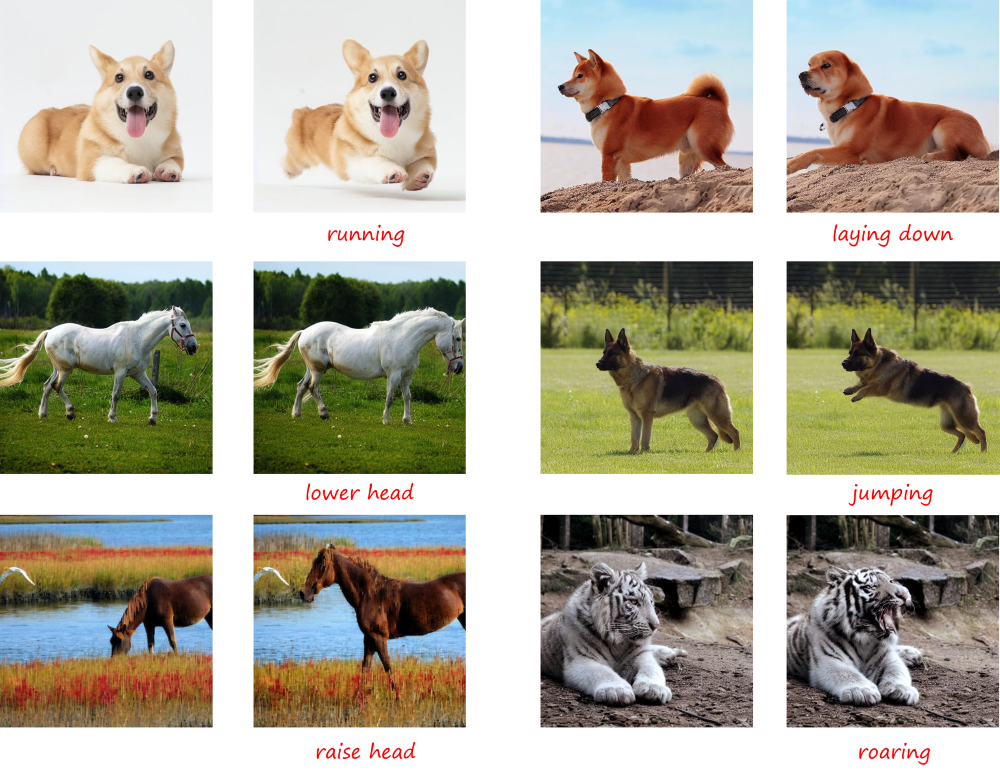}
\caption{\textbf{Creatures pose change.} We use verbs to manipulate the pose of creatures.}
\label{fig:pose_change}
\end{figure}

We showcase our method's ability to handle various editing tasks, such as object replacement, attribute manipulation, style transfer, and pose/shape changes, through more visual results. Additional results of object-level edits like replacement and attribute manipulation are provided in Fig. \ref{fig:object_replacement} and Fig. \ref{fig:attribute_change}. Fig. \ref{fig:style_transfer} shows the capability to transfer the image style with text guidance. Fig. \ref{fig:shape_change} and Fig. \ref{fig:pose_change} demonstrate that our method can handle complex non-rigid editing tasks such as changing the pose of creatures and the shape of items.

\section{Limitations and Discussion}
\label{sec:limitations}
Although our editing method has shown effectiveness in various types of editing tasks, we have identified some limitations that require further improvements. While our feature-sharing pipeline performs well in common scenarios, we have observed its weakness in the human face domain, especially when dealing with high-resolution and detailed face samples. We hypothesize that this is due to the incompatible expressiveness of low-resolution features of U-Net, resulting in the loss of high-density information from the source image during the editing process. Moreover, in some cases, our method can generate unreasonable representation that could be caused by ambiguous language descriptions like "a purple car" which may lead to the changed color mapped to the wheels. We leave these challenges as our future research we may extend our work to a more powerful generative prior like SDXL and enhance the feature interaction and optimization proficiency.

\begin{figure}[t!]
\centering
\includegraphics[width=1\textwidth]{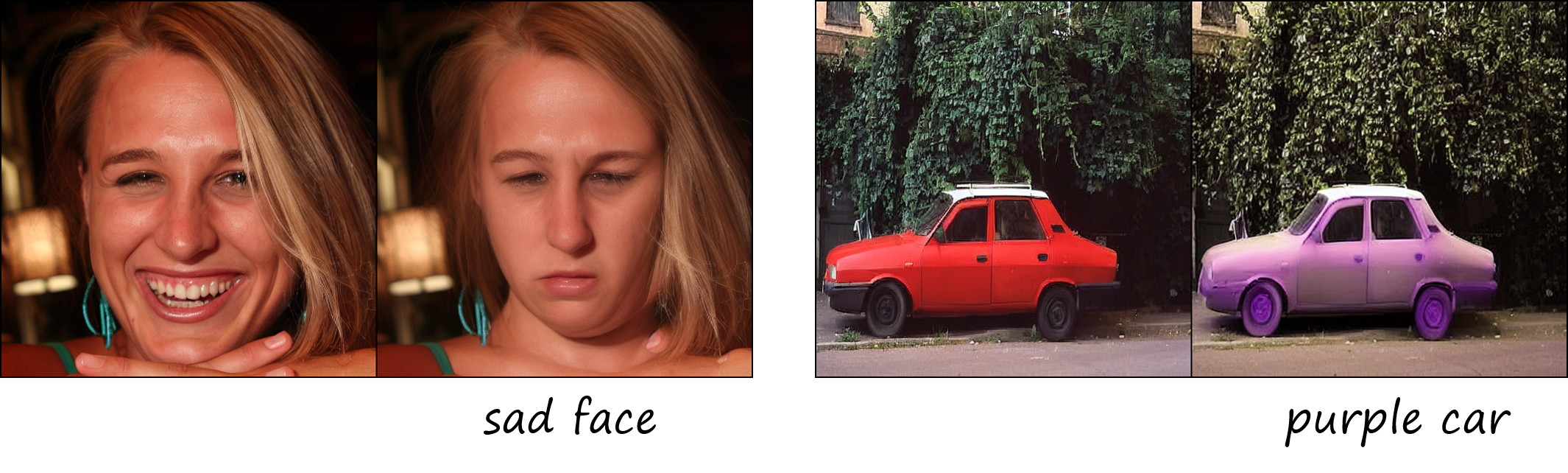}
\caption{\textbf{Limitations.} Our method, in some cases, sees limitations in the human face domain. It may generate unreasonable visual representations such as the colored wheel which could be rectified by using a more precise language description.}
\label{fig:limitations}
\end{figure}

\par\vfill\par

\end{document}